\documentclass[letterpaper]{article}
\usepackage{aaai}
\usepackage{times}
\usepackage{helvet}
\usepackage{courier}
\usepackage{amsmath}
\usepackage{amsfonts}
\usepackage{graphicx}
\usepackage{mathtools}
\usepackage{subcaption}
\usepackage{multicol}
\usepackage{caption}
\usepackage{natbib}
\usepackage{color}
\usepackage{placeins}
\nocopyright
 
\begin{document}
%
\title{Syntax-Infused Transformer and BERT models for \\ Machine Translation and Natural Language Understanding}
\author{Dhanasekar Sundararaman, Vivek Subramanian, Guoyin Wang, Shijing Si, \\ \textbf{\Large Dinghan Shen, Dong Wang, Lawrence Carin} \\
Department of Electrical \& Computer Engineering, Duke University\\
Durham, NC 27708\\
\{dhanasekar.sundararaman, vivek.subramanian, guoyin.wang, shijing.si, dinghan.shen, dong.wang363, lcarin\}@duke.edu\\
}
\maketitle
\begin{abstract}
\begin{quote}
Attention-based models have shown significant improvement over traditional algorithms in several NLP tasks. The Transformer, for instance, is an illustrative example that generates abstract representations of tokens inputted to an encoder based on their relationships to all tokens in a sequence. Recent studies have shown that although such models are capable of learning syntactic features purely by seeing examples, explicitly feeding this information to deep learning models can significantly enhance their performance. Leveraging syntactic information like part of speech (POS) may be particularly beneficial in limited training data settings for complex models such as the Transformer.  We show that the syntax-infused Transformer with multiple features achieves an improvement of 0.7 BLEU when trained on the full WMT '14 English to German translation dataset and a maximum improvement of 1.99 BLEU points when trained on a fraction of the dataset. In addition, we find that the incorporation of syntax into BERT fine-tuning outperforms baseline on a number of downstream tasks from the GLUE benchmark.
\end{quote}
\end{abstract}

\section{Introduction}
Attention-based deep learning models for natural language processing (NLP) have shown promise for a variety of machine translation and natural language understanding tasks. For word-level, sequence-to-sequence tasks such as translation, paraphrasing, and text summarization, attention-based models allow a single token ($e.g.$, a word or subword) in a sequence to be represented as a combination of all tokens in the sequence \citep{luong2015effective}. The distributed context allows attention-based models to infer rich representations for tokens, leading to more robust performance. One such model is the Transformer, which features a multi-headed self- and cross-attention mechanism that allows many different representations to be learned for a given token in parallel \citep{vaswani2017attention}. The encoder and decoder arms each contain several identical subunits that are chained together to learn embeddings for tokens in the source and target vocabularies.

Though the Transformer works well across a variety of different language pairs, such as (English, German) and (English, French), it consists of a large number of parameters and relies on a significant amount of data and extensive training to accurately pick up on syntactic and semantic relationships. Previous studies have shown that an NLP model's performance improves with the ability to learn underlying grammatical structure of a sentence \citep{kuncoro2018lstms, linzen2016assessing}.
In addition, it has been shown that simultaneously training models for machine translation, part of speech (POS) tagging, and named entity recognition provides a slight improvement over baseline on each task for small datasets \citep{niehues2017exploiting}. Inspired by these previous efforts, we propose to utilize the syntactic features that are inherent in natural language sequences, to enhance the performance of the Transformer model.

We suggest a modification to the embeddings fed into the Transformer architecture, that allows tokens inputted into the encoder to attend to not only other tokens but also syntactic features including POS, case, and subword position. These features are identified using a separate model (for POS) or are directly specified (for case and subword position) and are appended to the one-hot vector encoding for each token. Embeddings for the tokens and their features are learned jointly during the Transformer training process. As the embeddings are passed through the layers of the Transformer, the representation for each token is synthesized using a combination of word and syntactic features.

We evaluate the proposed model on  English to German (EN-DE) translation on the WMT '14 dataset. For the EN-DE translation task, we utilize multiple syntactic features including POS, case and subword tags that denote the relative position of subwords within a word \citep{sennrich2016linguistic}. Like POS, case is a categorical feature, which can allow the model to distinguish common words from important ones. Subword tags can help bring cohesion among subwords of a complex word (say, ``amalgamation'') so that their identity as a unit is not compromised by tokenization. We prove that the incorporation of these features improves the translation performance in the EN-DE task with a number of different experiments. We show that the BLEU score improvements of the feature-rich syntax-infused Transformer uniformly outperforms the baseline Transformer as a function of the training data size. Examining the attention weights learned by the proposed model further justifies the effectiveness of incorporating syntactic features.

We also experiment with this modification of embeddings on the BERT\textsubscript{BASE} model on a number of General Language Understanding Evaluation (GLUE) benchmarks and observe considerable improvement in performance on multiple tasks. With the addition of POS embeddings, the BERT\textsubscript{BASE + POS} model outperforms BERT\textsubscript{BASE} on 4 out of 8 downstream tasks.

To summarize, our main contributions are as follows:
\begin{enumerate}
\item We propose a modification to the trainable embeddings of the Transformer model, incorporating explicit syntax information, and demonstrate superior performance on EN-DE  machine translation task.
\item We modify pretrained BERT\textsubscript{BASE} embeddings by feeding in syntax information and find that the performance of BERT\textsubscript{BASE + POS} outperforms BERT\textsubscript{BASE} on a number of GLUE benchmark tasks.
\end{enumerate}

\section{Background}
\subsection{Baseline Transformer}

The Transformer consists of encoder and decoder modules, each containing several subunits that act sequentially to generate abstract representations for words in the source and target sequences \citep{vaswani2017attention}. As a preprocessing step, each word is first divided into subwords of length less than or equal to that of the original word \citep{sennrich2015neural}. These subwords are shared between the source and target vocabularies.

For all $m \in \{ 1,\: 2,\: \ldots, \: M\}$, where $M$ is the length of the source sequence, the encoder embedding layer first converts subwords $\mathbf{x}_m$ into embeddings $\mathbf{e}_m$:
\begin{align}
    \mathbf{e}_m = \mathbf{E}\mathbf{x}_m
\end{align}
where $\mathbf{E} \in \mathbb{R}^{D\times N}$ is a trainable matrix with column $m$ constituting the embedding for subword $m$, $N$ is the total number of subwords in the shared vocabulary, and $\mathbf{x}_m \in \{0, 1\}^N : \sum_i x_{mi} = 1$ is a one-hot vector corresponding to subword $m$. These embeddings are passed sequentially through six encoder subunits. Each of these subunits features a self-attention mechanism, that allows subwords in the input sequence to be represented as a combination of all subwords in the sequence. Attention is accomplished using three sets of weights: the key, query, and value matrices ($\mathbf{K}$, $\mathbf{Q}$, and $\mathbf{V}$, respectively). The key and query matrices interact to score each subword in relation to other subwords, and the value matrix gives the weights to which the score is applied to generate output embedding of a given subword. Stated mathematically,
\begin{align}
\begin{split}
    \mathbf{K} &= \mathbf{H}\mathbf{W}_K\\
    \mathbf{Q} &= \mathbf{H}\mathbf{W}_Q\\
    \mathbf{V} &= \mathbf{H}\mathbf{W}_V\\
    \mathbf{A} &= \mbox{softmax}\left(\frac{\mathbf{Q}\mathbf{K}^\top}{\sqrt{\rho}}\right)\mathbf{V}
\end{split}
\label{eq:attention}
\end{align}
where $\mathbf{H} = [\mathbf{h}_1 \: \mathbf{h}_2 \: \cdots \: \mathbf{h}_M]^\top \in \mathbb{R}^{M\times D}$ are the $D$-dimensional embeddings for a sequence of $M$ subwords indexed by $m$; $\mathbf{W}_K$, $\mathbf{W}_Q$, and $\mathbf{W}_V$ all $\in \mathbb{R}^{D\times P}$ are the projection matrices for keys, queries, and values, respectively; $\rho$ is a scaling constant (here, taken to be $P$) and $\mathbf{A} \in \mathbb{R}^{M \times P}$ is the attention-weighted representation of each subword. Note that these are subunit-specific -- a separate attention-weighted representation is generated by each subunit and passed on to the next. Moreover, for the first layer, $\mathbf{h}_m \coloneqq \mathbf{e}_m$.

The final subunit then passes its information to the decoder, that also consists of six identical subunits that behave similarly to those of the encoder. One key difference between the encoder and decoder is that the decoder not only features self-attention but also cross-attention; thus, when generating new words, the decoder pays attention to the entire input sequence as well as to previously decoded words.

\begin{figure}[t]
    \centering
    \includegraphics[width=0.45\textwidth]{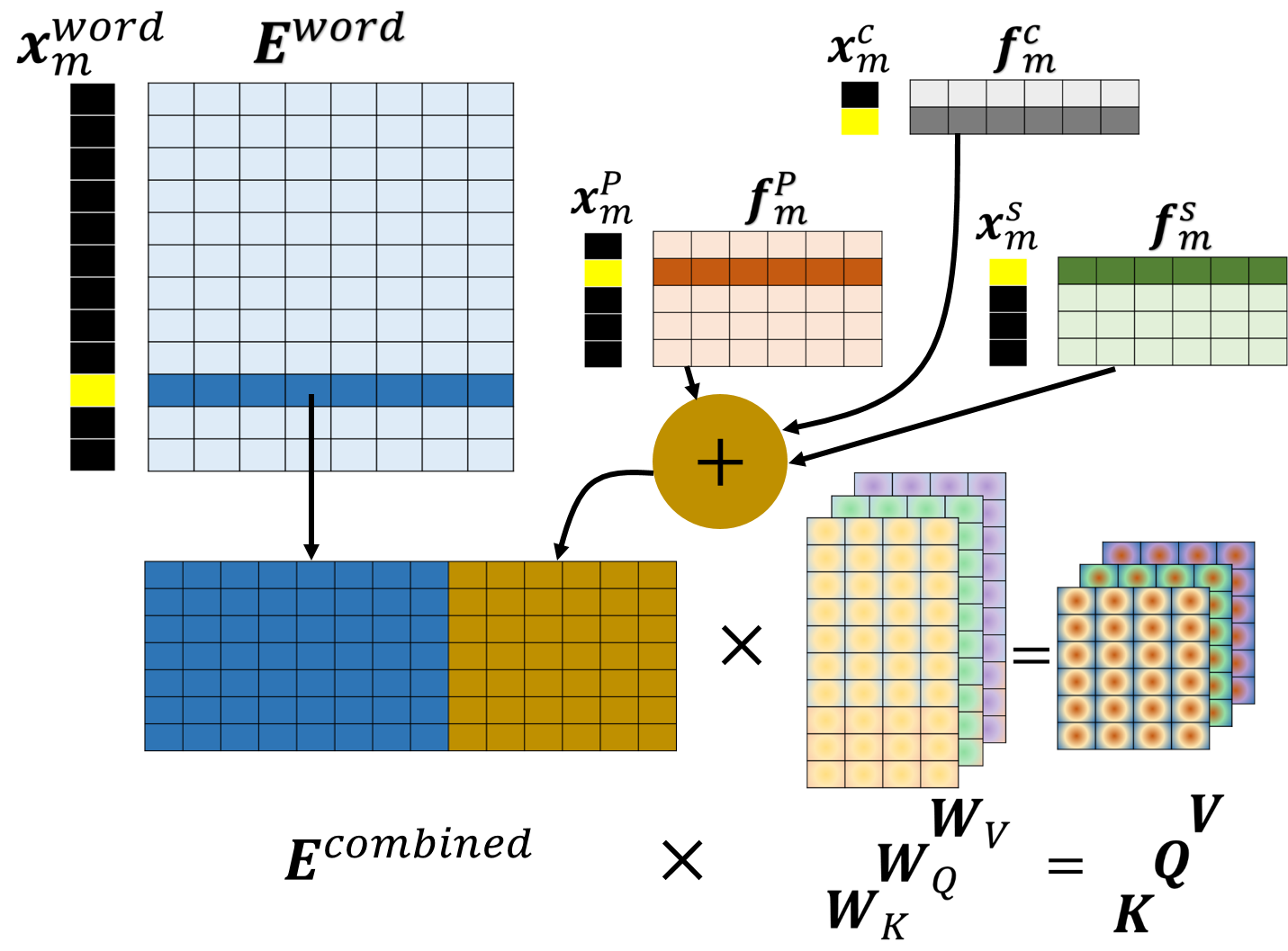}
    \caption{Formation of attention matrices ($\mathbf{K}$, $\mathbf{Q}$, and $\mathbf{V}$) with syntactic information. The left column shows the word embedding matrix; the embedding matrices for the various features are shown on top. 
    Embeddings for the chosen features are either concatenated or summed together (denoted by $\oplus$) and finally, concatenated to the word embeddings. Matrix multiplication with learned weights results in $\mathbf{K}$, $\mathbf{Q}$, and $\mathbf{V}$.
    The attention matrices are double shaded to indicate the mix of word and syntax information.}
    \label{fig:subwordconcat}
\end{figure}

\subsection{BERT}
While the Transformer is able to generate rich representations of words in a sequence by utilizing attention, its decoder arm restricts it to be task-specific. The word embeddings learned by the Transformer encoder, however, can be fine-tuned to perform a number of different downstream tasks. Bidirectional encoder representations of Transformers (BERT) is an extension of the Transformer model that allows for such fine-tuning. The BERT model is essentially a Transformer encoder (with number of layers $l$, embedding dimension $D$, and number of attention heads $\alpha$) which is pre-trained using two methods: masked language modeling (MLM) and next-sentence prediction (NSP). Subsequently, a softmax layer is added, allowing the model to perform various tasks such as classification, sequence labeling, question answering, and language inference. According to \citep{devlin2018bert}, BERT significantly outperforms previous state-of-the-art models on the eleven NLP tasks in the GLUE benchmark \citep{wang2018glue}.

\section{Model}
\subsection{Syntax-infused Transformer}
Syntax is an essential feature of grammar that facilitates generation of coherent sentences. For instance, POS dictates how words relate to one another ($e.g.$, verbs represent the actions of nouns, adjectives describe nouns, etc.). Studies have shown that when trained for a sufficiently large number of steps, NLP models can potentially learn underlying patterns about text like syntax and semantics, but this knowledge is imperfect \citep{jawahar2019does}. However, works such as \citep{kuncoro2018lstms, linzen2016assessing} show that NLP models that acquire even a weak understanding of syntactic structure through training demonstrate improved performance relative to baseline. Hence, we hypothesize that explicit prior knowledge of syntactic information can benefit NLP models in a variety of tasks.

To aid the Transformer in more rapidly acquiring and utilizing syntactic information for better translation, we ($i$) employ a pretrained model\footnote{https://spacy.io/} to tag words in the source sequence with their POS, ($ii$) identify the case of each word, and ($iii$) identify the position of each subword relative to other subwords that are part of the same word (subword tagging). We then append trainable syntax embedding vectors to the token embeddings, resulting in a combined representation of syntactic and semantic elements.

Specifically, each word in the source sequence is first associated with its POS label according to syntactic structure. After breaking up words into their corresponding subwords (interchangeably denoted as tokens), we assign each subword the POS label of the word from which it originated. For example, if the word \texttt{sunshine} is broken up into subwords \texttt{sun}, \texttt{sh}, and \texttt{ine}, each subword would be assigned the POS \texttt{NOUN}. The POS embeddings are then extracted from a trainable embedding matrix using a look-up table, in a manner similar to that of the subword embeddings (see Figure \ref{fig:subwordconcat}). The POS embeddings $\mathbf{f}_m^{P}$ of each subword (indexed by $m$) are then concatenated with the subword embeddings $\mathbf{e}_m \in \mathbb{R}^{D-d}$ to create a combined embedding where $d$ is the dimension of the feature embedding.

\begin{figure}[t]
    \centering
    \includegraphics[width=0.47\textwidth]{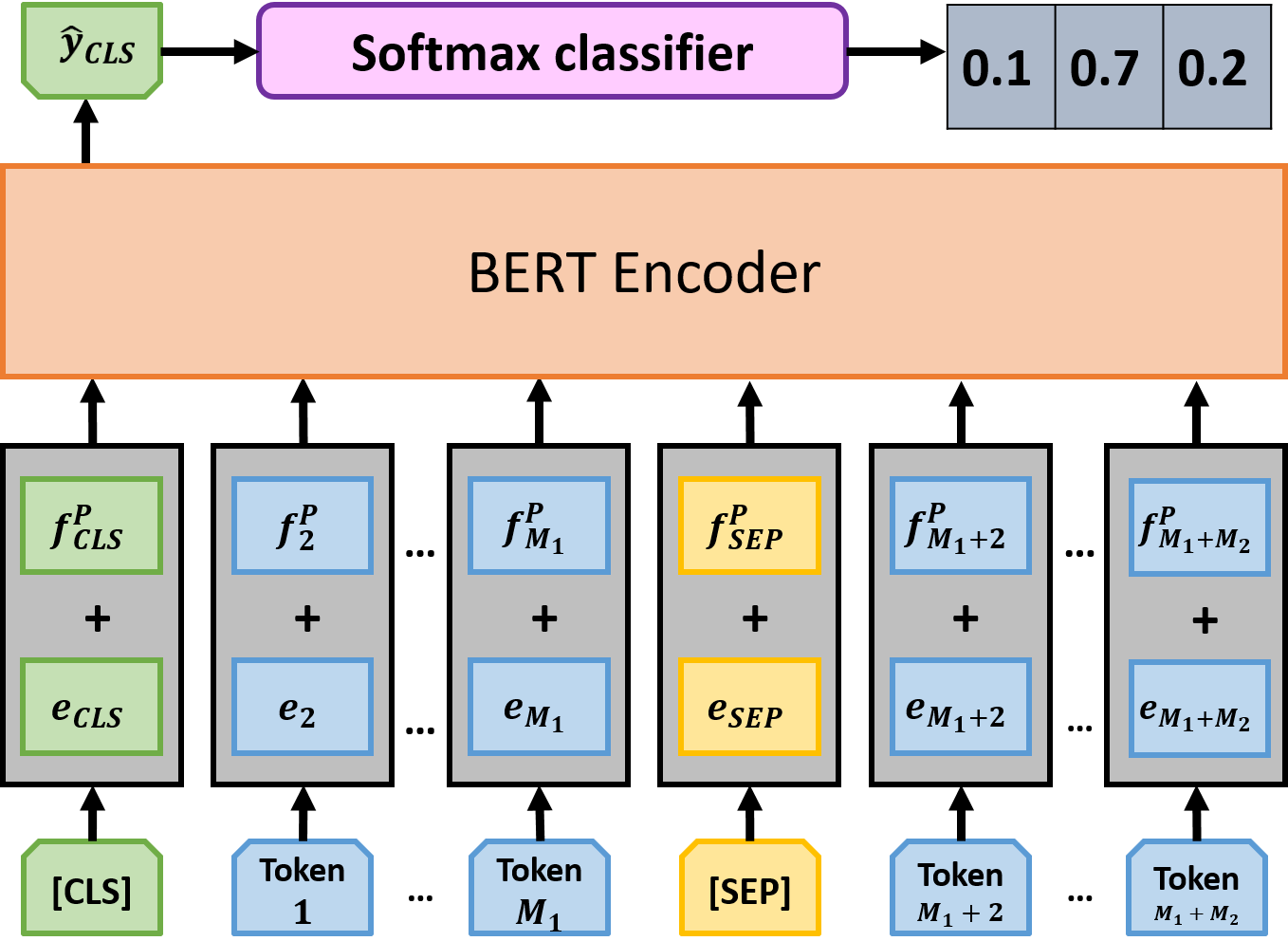}
    \caption{The BERT\textsubscript{BASE + POS} model. Token embeddings are combined with trainable POS embeddings and fed into the BERT encoder. The final embedding of the [CLS] token is fed into a softmax classifer for downstream classification tasks. The model is illustrated as taking in a pair of sequences but single sequence classification is also possible.}
    \label{fig:bertpos}
\end{figure}

In a similar manner, we incorporate case and subword position features. For case, we use a binary element $z_m^c \in \{0, 1\}$ to look up a feature embedding $\mathbf{f}_m^c$ for each subword, depending on whether the original word is capitalized. For subword position, we use a categorical element $z_m^s \in \{B, M, E, O\}$ to identify a feature embedding $\mathbf{f}_m^s$ for each subword depending on whether the subword is at the beginning ($B$), middle ($M$), or end ($E$) of the word; if the subword comprises the full word, it is given a tag of $O$. These are then added onto the POS embedding. Mathematically, in the input stage, $\mathbf{h}_m$ becomes: 
$$[\mathbf{e}_m^\top \: \mathbf{f}_m^\top]^\top = \mathbf{h}_m' \in \mathbb{R}^{D}$$ 
where $\mathbf{f}_m = \mathbf{f}_m^{P} \oplus \mathbf{f}_m^{c} \oplus \mathbf{f}_m^{s} \in \mathbb{R}^d$ is the learned embedding for the syntactic features of subword $m$ in the sequence of $M$ subwords and $\oplus$ denotes either the concatenation or summation operation.

We conjecture that our syntax-infused Transformer model can boost translation performance by injecting grammatical relationships, without having to learn them from examples.

\subsection{Syntax-infused BERT}
Adding syntactic features to the BERT model is a natural extension of the above modification to the Transformer. As mentioned above, embeddings trained by BERT can be utilized for a variety of downstream tasks. We hypothesize that infusing BERT with syntactic features is beneficial in many of these tasks, especially those involving semantic structure.

Many of the datasets on which we evaluate our modified BERT model are low-resource (as few as 2.5k sentences) relative to those on which we evaluate the syntax-infused Transformer; hence, we choose to utilize only POS as a syntactic feature for BERT. We consider two approaches for combining POS features with the pre-trained embeddings in BERT, a model we denote as BERT\textsubscript{BASE + POS}: (1) addition of the trainable POS embedding vector of dimension $d=D$ to the token embedding and (2) concatenation of the POS embedding with the token embedding. To make a fair comparison with BERT\textsubscript{BASE}, the input dimension $D$ of the encoder must match that of BERT\textsubscript{BASE} ($D=768$). Thus, if option 2 is used, the concatenated embedding must be passed through a trainable affine transformation with weight matrix of size $(D+d) \times D$ . While this option provides a more robust way to merge POS and word embeddings, it requires learning a large matrix, which is problematic for downstream tasks with very little training data. Hence, to facilitate training for these tasks and to standardize the comparison across different downstream tasks, we choose to use the first approach. Therefore, for a given token, its input representation is constructed by summing the corresponding BERT token embeddings with POS embeddings (see Figure \ref{fig:bertpos}).

Mathematically, the input tokens $\mathbf{h}_m' \in \mathbb{R}^D$ are given by $\mathbf{h}_m' = \mathbf{e}_m + \mathbf{f}_m^P$, where $\mathbf{e}_m$ is the BERT token embedding and $\mathbf{f}_m^P$ is the POS embedding for token $m$. For single sequence tasks, $m = 1, 2, \ldots, M$, where $M$ is the number of tokens in the sequence; while for paired sequence tasks, $m = 1, 2, \ldots, M_1 + M_2$, where $M_1$ and $M_2$ are the number of tokens in each sequence. As is standard with BERT, for downstream classification tasks, the final embedded representation $\mathbf{\hat{y}}_{CLS}$ of the first token (denoted as [CLS]) is passed through a softmax classifer to generate a label.

\section{Datasets and Experimental Details}
For translation, we consider WMT '14 EN-DE dataset. The WMT '14 dataset consists of 4.5M training sentences. Validation is performed on newstest2013 (3000 sentences) and testing is on the newstest2014 dataset (2737 sentences, \citep{zhang2019improving}). Parsers that infer syntax from EN sentences are typically trained on a greater number and variety of sentences and are therefore more robust than parsers for other languages. Since one of the key features of our models is to incorporate POS features into the source sequence, we translate \textit{from} EN \textit{to} DE. While incorporating all linguistic features described above is generally beneficial to NLP models, adding features may compromise the model by restricting the number of dimensions allocated to word embeddings, which still the play the primary role. We consider this tradeoff in greater detail below.

\subsection{Machine translation}

We train both the baseline and syntax-infused  Transformer for 100,000 steps. All hyperparameter settings of the baseline Transformer, including embedding dimensions of the encoder and decoder, match those of \citep{vaswani2017attention}. We train the syntax-infused Transformer model using 512-dimensional embedding vectors. In the encoder, $D = 492$ dimensions are allocated for word embeddings while $d = 20$ for feature embeddings (chosen by hyperparameter tuning). In the decoder, all 512 dimensions are used for word embeddings (since we are interested only in decoding words, not word-POS pairs). 

The model architecture consists of six encoder and six decoder layers, with eight heads for multi-headed attention. Parameters are initialized with Glorot \citep{glorot2010understanding}. We use a dropout rate of 0.1 and batch size of 4096. We utilize the Adam optimizer to train the model with $\beta_1 = 0.9$ and $\beta_2 = 0.998$; gradients are accumulated for two batches before updating parameters. A label-smoothing factor of 0.1 is employed.

\begin{table}[t]
\centering
\begin{tabular}{|c|c|c|c|}
\hline
Data  &  Number of & Baseline  & Syntax-infused \\
Fraction & Sentences & Transformer &Transformer \\
\hline
1\% & 45k  & 1.10 & \textbf{1.67}  \\ \hline
5\% & 225k & 8.51 & \textbf{10.50}  \\ \hline
10\%& 450k & 16.28 & \textbf{17.28} \\ \hline
25\%& 1.1M & 22.72 & \textbf{23.24} \\ \hline
50\%& 2.25M& 25.41 & \textbf{25.74} \\ \hline
100\%&4.5M & 28.94 & \textbf{29.64} \\    \hline
\end{tabular}
\caption{BLEU scores for different proportions of the data for baseline Transformer vs syntax-infused Transformer for the EN-DE task on newstest2014.}
\label{tab:hrmtvsdatasize}
\end{table}

\begin{figure}[t]
\centering
    \begin{subfigure}{0.23\textwidth}
        \centering
        \includegraphics[width=\textwidth, height=\textwidth]{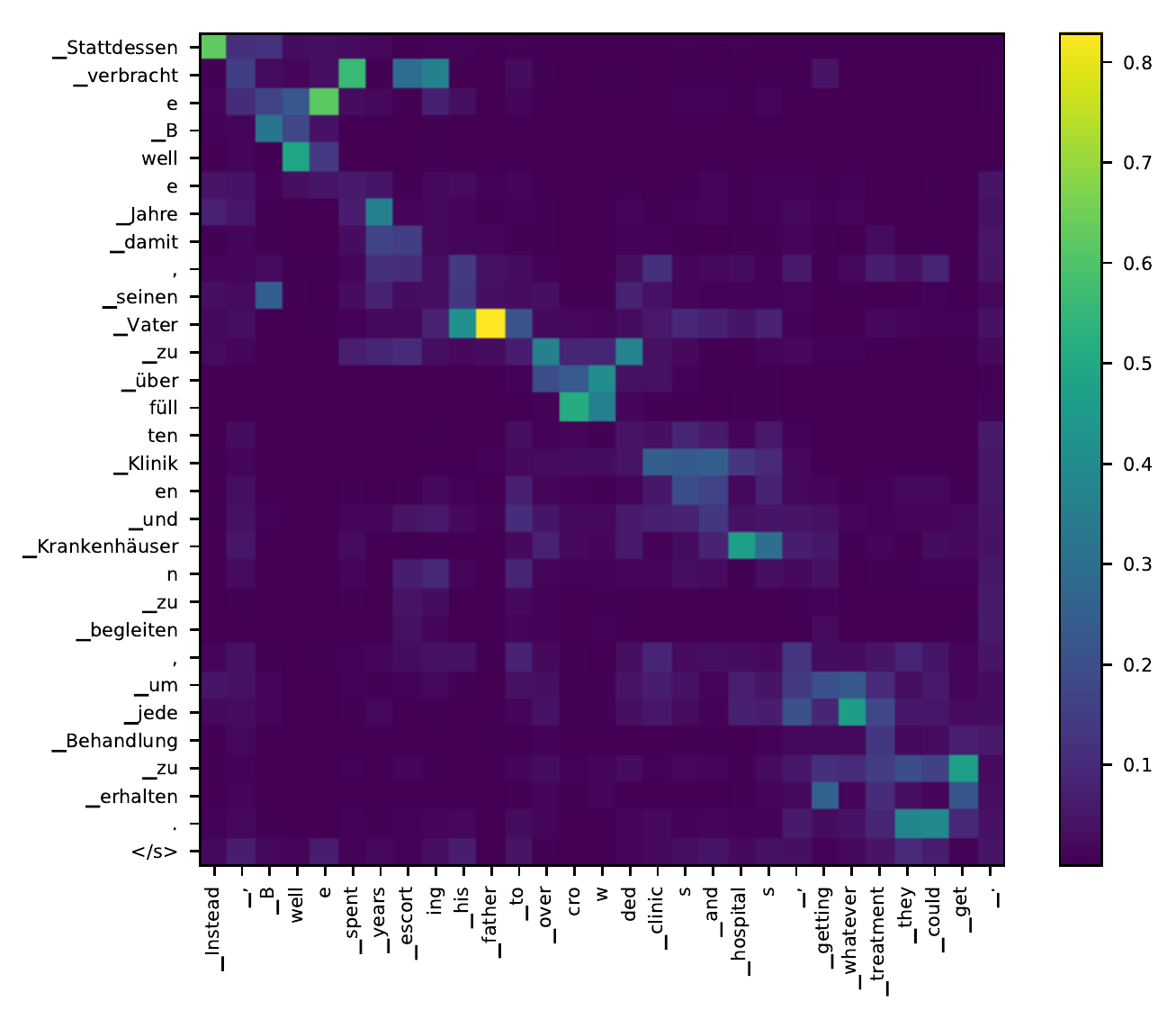}
        \caption{Baseline (EN-DE)}
    \end{subfigure}
    \begin{subfigure}{0.225\textwidth}
        \centering
        \includegraphics[width=\textwidth, height=\textwidth]{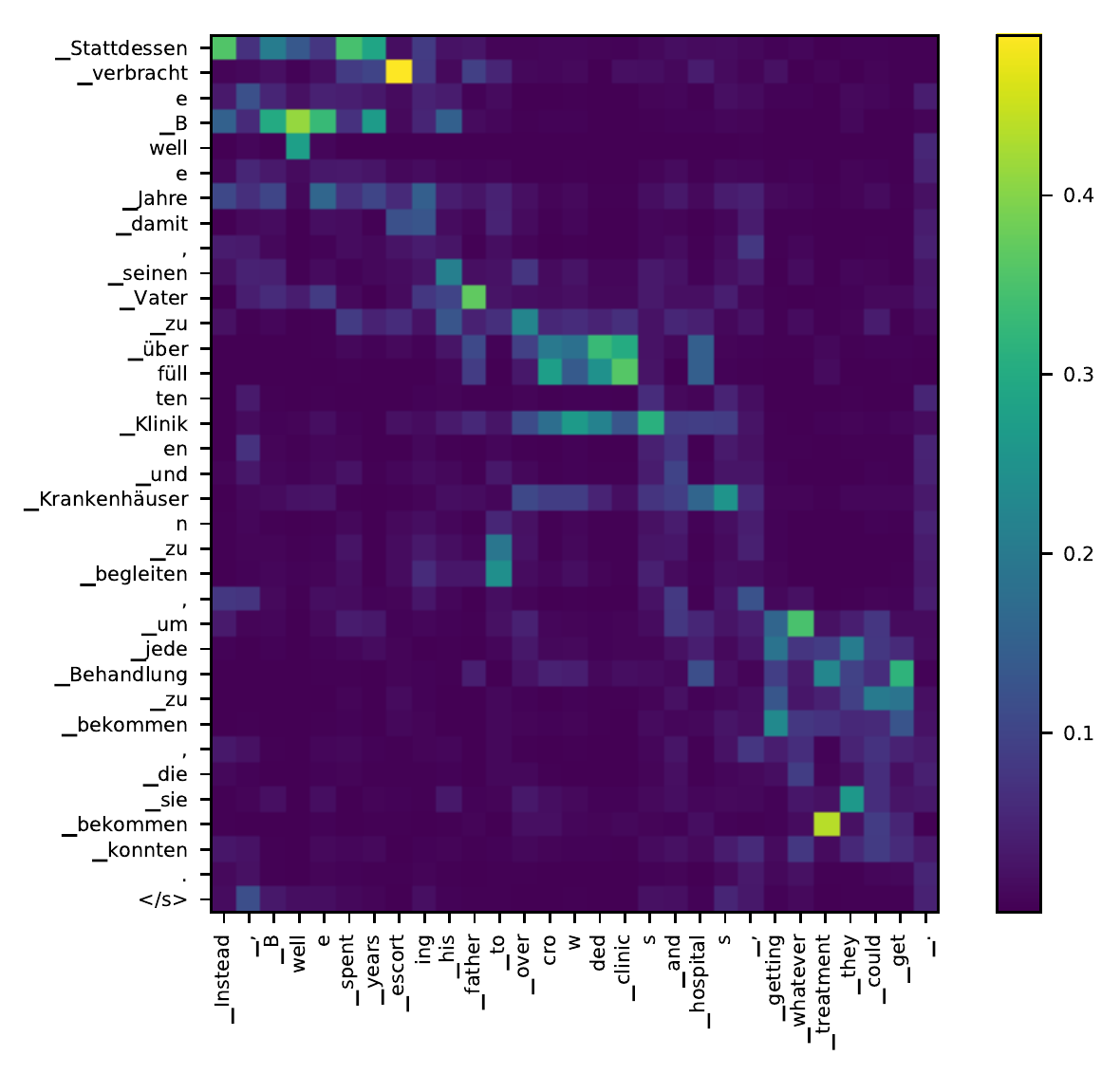}
        \caption{Syntax-infused (EN-DE)}
    \end{subfigure}
    \caption{Comparison of attention for example sentences translated by baseline and POS Transformer models (obtained from the last layer). Rows depict the attention score for a given target subword to each of the subwords in the source sequence. In syntax-infused models for EN-DE translation, we find that attention is more widely distributed across subwords. For instance, the subword ``Vater'' (the German word for ``father'') attends mostly to the nearby subwords ``his'' and ``father'' in the base model while ``Vater'' also attends to the more distant words ``Bwelle'' (a person) and ``escorting'' in the syntax-infused model. This suggests that the syntax-infused model is able to better connect disparate parts of a sentence to aid translation. Note that the number of rows in the baseline and syntax-infused Transformer are different because each produces different predictions.}
    \label{fig:attention}
\end{figure}

 The context and size of the EN-DE translation dataset is quite different compared that of the datasets on which POS tagging methods are typically trained, implying that the POS tagging model may not generalize well. Hence, we include not only POS but also case and subword tag features. The training procedure is identical to that of \citep{vaswani2017attention} except that, for the syntax-infused Transformer, the dimension $d$ of features $\mathbf{f}_m$ is chosen to be 20 by doing a grid search over the range of 8 to 64.

\subsection{Natural language understanding}
The General Language Understanding Evaluation (GLUE) benchmark \citep{wang2018glue} is a collection of different natural language understanding tasks evaluated on eight datasets: Multi-Genre Natural Language Inference (MNLI), Quora Question Pairs (QQP), Question Natural Language Inference (QNLI), Stanford Sentiment Treebank (SST-2), The Corpus of Linguistic Acceptability (CoLA), The Semantic Textual Similarity Benchmark (STS-B), Microsoft Research Paraphrase Corpus (MRPC), and Recognizing Textual Entailment (RTE). For a summary of these datasets, see \citep{devlin2018bert}. We use POS as the syntactic feature for BERT for these tasks. Aside from the learning rate, we use identical hyperparameter settings to fine-tune both the BERT\textsubscript{BASE} and BERT\textsubscript{BASE + POS} models for each task. This includes a batch size of 32 and 3 epochs of training for all tasks. For each model, we also choose a task-specific learning rate among the values $\{5, 4, 3, 2\} \times 10^{-5}$, which is standard for BERT\textsubscript{BASE}.




\begin{table*}[!h]
\begin{center}
\begin{tabular}{|c|c|c|c|c|c|c|c|c|c|}
\hline \bf System &  \bf MNLI &  \bf QQP &  \bf QNLI &  \bf SST-2 &  \bf CoLA &  \bf STS-B &  \bf MRPC &  \bf RTE & \bf Average \\ 

& 392k & 363k & 108k & 67k & 8.5k & 5.7k & 3.5k & 2.5k & - \\ \hline

Pre-OpenAI SOTA & 80.6/80.1 & 66.1 & 82.3 & 93.2 & 35.0 & 81.0 & 86.0 & 61.7 & 74.0 \\ \hline

BiLSTM+ELMo+Attn & 76.4/76.1 & 64.8 & 79.8 & 90.4 & 36.0 & 73.3 & 84.9 & 56.8 & 71.0 \\ \hline

OpenAI GPT & 82.1/81.4 & 70.3 & 87.4 & 91.3 & 45.4 & 80.0 & 82.3 & 56.0 & 75.1 \\ \hline

BERT\textsubscript{BASE} & 84.6/83.4 & 71.2 & 90.5 & 93.5 & 52.1 & 85.8 & 88.9 & 66.4 & 79.6 \\ \hline

BERT\textsubscript{BASE + POS} & 84.4/83.3 & \textbf{71.4} & 90.4 & \textbf{93.9} & \textbf{52.9} & 85.5 & 88.8 & \textbf{66.9} & \textbf{79.7} \\ \hline

\hline
\end{tabular}
\end{center}
\caption{GLUE test results scored using the GLUE evaluation server. The number below each task denotes the number of training examples. The scores in bold denote the tasks for which BERT\textsubscript{BASE + POS} outperforms BERT\textsubscript{BASE}.}
\label{tab:glueresults}
\end{table*}

\section{Experimental Results}

\subsection{Machine translation}
We evaluate the impact of infusing syntax into the baseline Transformer for the EN-DE translation task. We add three features namely POS, subword tags, and case to aid Transformer model learn underlying patterns about the sentences.

With more than one feature, there are multiple ways to incorporate feature embeddings into the word embeddings. For a fair comparison to the Transformer baseline, we use a total of 512 dimensions for representing both the word embeddings as well as feature embeddings. One important tradeoff is that as the dimensionality of the syntax information increases, the dimensionality for actual word embeddings decreases. Since POS, case, and subword tags have only a limited number of values they can take, dedicating a high dimensionality for each feature proves detrimental (experimentally found). We find that the total feature dimension for which the gain in BLEU score is maximized is 20 (found through grid search). This means that (1) each feature embedding dimension can be allocated to 20 and summed together or (2) the feature embeddings can be concatenated to each other such that their total dimensionality is 20. Therefore, in order to efficiently learn the feature embeddings while also not sacrificing the word embedding dimensionality, we find that summing the embeddings for all three different features of $d=20$ and concatenating the sum to the word embeddings of $D=492$ gives the maximum performance on translation. We also find that incorporation of a combination of two features among \{POS, case, subword tags\} does not perform as well as having all the three features.

In Table \ref{tab:hrmtvsdatasize}, we vary the proportion of data used for training and observe the performance of both the baseline and syntax-infused Transformer. The syntax-infused model markedly outperforms the baseline model, offering an improvement of 0.57, 1.99, 1, 0.52, 0.33, and 0.7 points, respectively, for 1, 5, 10, 25, 50, and 100\% of the data. It is notable that the syntax-infused model translates the best relative to the baseline when only a fraction of the dataset is used for training.  Specifically, the maximum improvement is 1.99 BLEU points when only 10\% of the training data is used. This shows that explicit syntax information is most helpful under limited training data conditions. As shown in Figure \ref{fig:attention}(a)-(b), the syntax-infused model is better able to capture connections between tokens that are far apart yet semantically related, resulting in improved translation performance. In addition, Table \ref{tab:translationexamples} shows a set of sample German predictions made by the baseline and syntax-infused Transformer. 

\begin{table*}[t]
\begin{center}
\begin{tabular}{|p{5.15cm}|p{5.15cm}|p{5.15cm}|}
\hline \bf Reference &  \bf Baseline Transformer &  \bf Syntax-infused Transformer  \\ 

\hline  Parken in Frankfurt k{\"o}nnte bald empfindlich teurer werden . &  Das Personal war sehr freundlich und hilfsbereit . & \textcolor{blue}{Parken in Frankfurt} \textcolor{blue}{k{\"o}nnte bald} sp{\"u}rbar \textcolor{blue}{teurer} sein .
 \\ \hline
 
Die zur{\"u}ckgerufenen Modelle wurden zwischen dem 1. August und 10. September hergestellt .
&
Zwischen August 1 und September 10.
&
Die \textcolor{blue}{zur{\"u}ckgerufenen Modelle wurden zwischen dem 1. August und 10. September} gebaut
\\ \hline

Stattdessen verbrachte Bwelle Jahre damit , seinen Vater in {\"u}berf{\"u}llte Kliniken und Hospit{\"a}ler zu begleiten , um dort die Behandlung zu bekommen , die sie zu bieten hatten .
&
Stattdessen verbrachte Bwelle Jahre damit , seinen Vater mit {\"u}ber f{\"u}llten Kliniken und Krankenh{\"q}usern zu beherbergen .
&
Stattdessen verbrachte Bwelle Jahre damit , seinen Vater zu {\"u}berf{\"u}llten Kliniken und Krankenh{\"a}usern zu begleiten , \textcolor{blue}{um} jede \textcolor{blue}{Behandlung zu bekommen , die sie bekommen} konnten .
\\ \hline

Patek kann gegen sein Urteil noch Berufung ein legen .
&
Patek kann noch seinen Satz an rufen .
&
Patek mag sein \textcolor{blue}{Urteil noch Berufung ein legen} .
\\ \hline
\end{tabular}
\end{center}
\caption{Translation examples of baseline Transformer vs. syntax-infused Transformer on the EN-DE dataset. The text highlighted in blue represents words correctly predicted by the syntax-infused model but not by the baseline Transformer.}
\label{tab:translationexamples}
\end{table*}

\begin{table*}[t]
\begin{center}
\begin{tabular}{|p{8.5cm}|p{4.8cm}|p{2.15cm}|}
\hline \bf Sentence 1 &  \bf Sentence 2 &  \bf True label  \\ \hline 
The Qin (from which the name China is derived) established the approximate boundaries and basic administrative system that all subsequent dynasties were to follow . & Qin Shi Huang was the first Chinese Emperor . & Not entailment \\ \hline
In Nigeria, by far the most populous country in sub-Saharan Africa, over 2.7 million people are infected with HIV . & 2.7 percent of the people infected with HIV live in Africa . & Not entailment \\ \hline 
\end{tabular}
\end{center}
\caption{Examples of randomly chosen sentences from the RTE dataset (for evaluation of entailment between pairs of sentences) that were  misclassified by BERT\textsubscript{BASE} and correctly classified by BERT\textsubscript{BASE + POS}.}
\label{tab:bertexamples}
\end{table*}

\subsection{Natural language understanding}
The results obtained for the BERT\textsubscript{BASE + POS} model on the GLUE benchmark test set are presented in Table \ref{tab:glueresults}. BERT\textsubscript{BASE + POS} outperforms BERT\textsubscript{BASE} on 4 out of the 8 tasks. The improvements range from marginal to  significant, with a maximum improvement of 0.8 points of the POS model over BERT\textsubscript{BASE} on CoLA. Fittingly, CoLA is a task which assesses the linguistic structure of a sentence, which is explictly informed by POS embeddings. Moreover, BERT\textsubscript{BASE + POS} outperforms BERT\textsubscript{BASE} on tasks that are concerned with evaluating semantic relatedness. For examples of predictions made on the RTE dataset, see Table \ref{tab:bertexamples}.

\section{Related Works}

Previous work has sought to improve the self-attention module to aid NLP models. For instance, \citep{yang2018modeling} introduced a Gaussian bias to model locality, to enhance model ability to capture local context while also maintaining the long-range dependency. Instead of absolute positional embeddings, \citep{shaw2018self} experimented with relative positional embeddings or distance between sequences and found that it led to a drastic improvement in performance.

Adding linguistic structure to models like the Transformer can be thought of as a way of improving the attention mechanism. The POS and subword tags act as a form of relative positional embedding by enforcing the sentence structure. \citep{li2018multi} encourages different attention heads to learn about different information like position and representation by introducing a disagreement regularization. In order to model the local dependency between words more efficiently, \citep{im2017distance} introduced distance between words and incorporated that into the self-attention.

Previous literature also has sought to incorporate syntax into deep learning NLP models. \citep{bastings2017graph} used syntax dependency tree information on a bidirectional RNN on translation systems by modeling the trees using Graph Convolutional Networks (GCNs) \citep{kipf2016semi}. Modeling source label syntax information has helped significantly in the Chinese-English translation \citep{li2017modeling} by linearizing parse trees to obtain drastic performance improvements. Adding a syntax-based distance constraint on the attention module, to generate a more semantic context vector, has proven to work for translation systems in the Chinese-English as well as English-German tasks.

These works affirm that adding syntax information can help the NLP models to translate better from one language to another and also achieve better performance measures.

\section{Conclusions}
We have augmented the Transformer network with syntax information for machine translation. The syntax-infused Transformer improvements were highest when a subset of the training data is used. We then distinguish the syntax-infused and baseline Transformer models by providing an interpretation of attention visualization. Additionally, we find that the syntax-infused BERT model performs better than baseline on a number of GLUE downstream tasks.

It is an open question whether the efficiency of these sophisticated models can further be improved by creating an architecture that is enabled to model the language structure more inherently than using end to end models. Future work may extend toward this direction.

\bibliography{aaai}
\bibliographystyle{aaai}

\end{document}